\documentclass[10pt,twocolumn,letterpaper]{article}

\usepackage{cvpr}
\usepackage{times}
\usepackage{epsfig}
\usepackage{graphicx}
\usepackage{amsmath}
\usepackage{amssymb}
\usepackage{array}
\usepackage{enumitem}
\usepackage{caption}
\usepackage[footnotesize]{subfigure}

\usepackage[lined,linesnumbered,ruled]{algorithm2e}

\SetCommentSty{mycommfont}

\DeclareMathOperator*{\argmin}{arg\,min}

\usepackage{multirow}
\usepackage{bbm}
\usepackage[pagebackref=true,breaklinks=true,letterpaper=true,colorlinks,bookmarks=false]{hyperref}

\newcommand\blfootnote[1]{%
  \begingroup
  \renewcommand\thefootnote{}\footnote{#1}%
  \addtocounter{footnote}{-1}%
  \endgroup
}

\cvprfinalcopy 

\def\cvprPaperID{2517} 

\ifcvprfinal\fi
\begin{document}


\title{Zero-Shot Visual Recognition using Semantics-Preserving \\ Adversarial Embedding Networks}

\author{Long Chen$^{1}$ \quad Hanwang Zhang$^{2}$ \quad Jun Xiao$^{1\ast}$ \quad Wei Liu$^{3}$ \quad Shih-Fu Chang$^{4}$ \\
$^{1}$Zhejiang University ~ $^2$Nanyang Technological University ~ $^{3}$Tencent AI Lab ~ $^{4}$Columbia University \\
{\tt\small \{longc,~junx\}@zju.edu.cn; hanwangzhang@ntu.edu.sg; \{wliu,~sfchang\}@ee.columbia.edu }\\
}

\thispagestyle{empty}
\twocolumn[{%
\maketitle
\thispagestyle{empty}
\begin{center}
    \centering
    \includegraphics[width=1\linewidth]{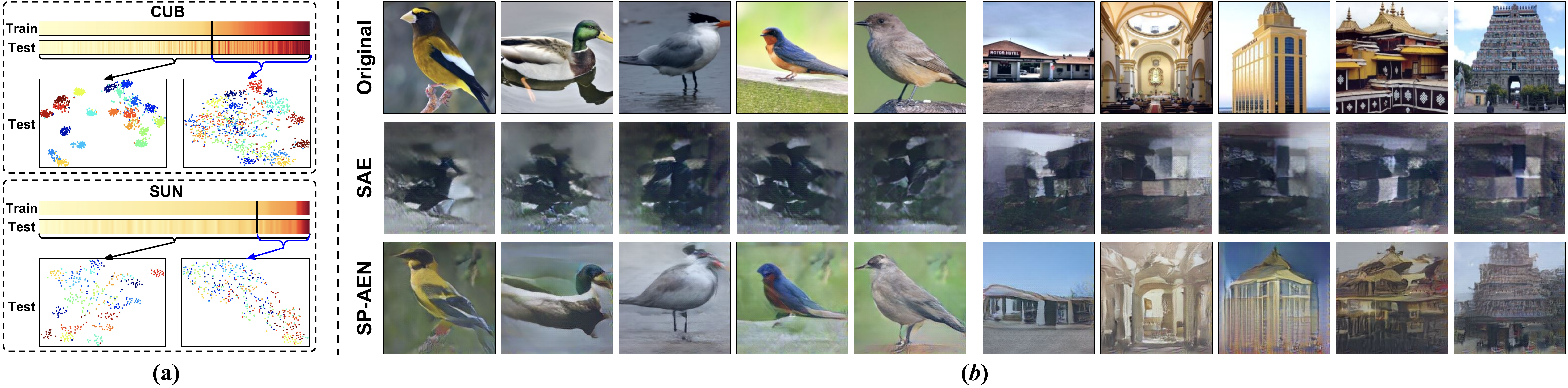}
    \captionof{figure}{{{(a) Attribute variance heat maps of the 312 attributes in CUB birds~\cite{welinder2010caltech} and the 102 attributes in SUN scenes~\cite{patterson2012sun} (lighter color indicates lower variance, \ie, lower discriminability) and the t-SNE~\cite{maaten2008visualizing} visualizations of the test images represented by all attributes (left) and only the high-variance ones (right). Some of the low-variance attributes (the lighter part to the left of the cut-off line) discarded at training are still needed in discriminating unseen test classes. (b) Comparison of reconstructed images using SAE~\cite{kodirov2017semantic} and our proposed SP-AEN method, which is shown to retain sufficient semantics for photo-realistic reconstruction.}}}
\label{fig:1}
\end{center}%
}]


\begin{abstract}
\blfootnote{$^\ast$Corresponding Author}We propose a novel framework called Semantics-Preserving Adversarial Embedding Network (SP-AEN) for zero-shot visual recognition (ZSL), where test images and their classes are both unseen during training. SP-AEN aims to tackle the inherent problem --- semantic loss --- in the prevailing family of embedding-based ZSL, where some semantics would be discarded during training if they are non-discriminative for training classes, but could become critical for recognizing test classes. Specifically, SP-AEN prevents the semantic loss by introducing an independent visual-to-semantic space embedder which disentangles the semantic space into two subspaces for the two arguably conflicting objectives: classification and reconstruction. Through adversarial learning of the two subspaces, SP-AEN can transfer the semantics from the reconstructive subspace to the discriminative one, accomplishing the improved zero-shot recognition of unseen classes. Comparing with prior works, SP-AEN can not only improve classification but also generate photo-realistic images, demonstrating the effectiveness of semantic preservation. On four popular benchmarks: CUB, AWA, SUN and aPY, SP-AEN considerably outperforms other state-of-the-art methods by an absolute performance difference of 12.2\%, 9.3\%, 4.0\%, and 3.6\% in terms of harmonic mean values~\cite{xian2017zero}.
\end{abstract}

\vspace{-0.5cm}
\section{Introduction}\label{sec:intro}
\begin{figure}[t]
	\centering
	\includegraphics[width=1\linewidth]{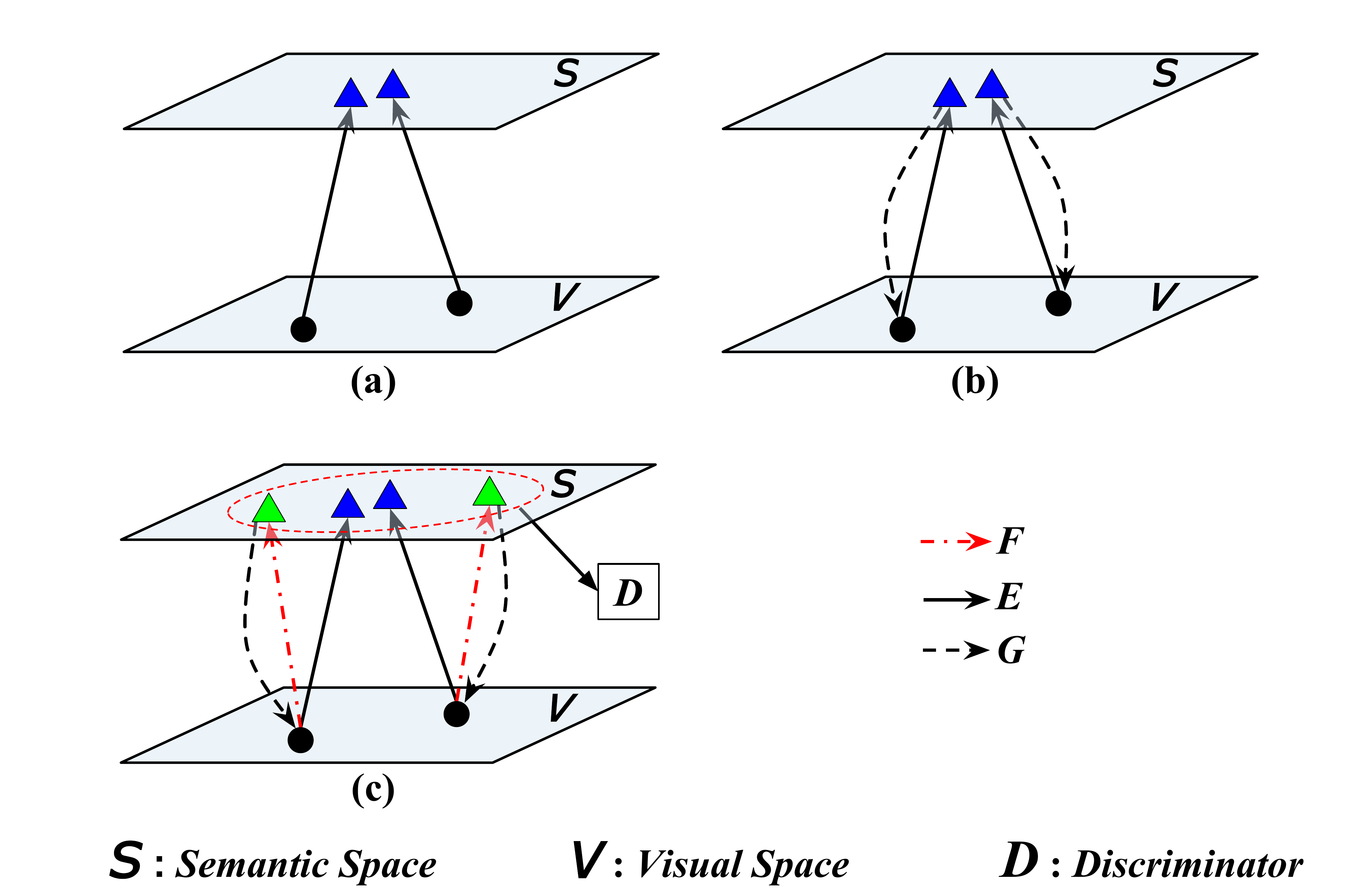}
	\caption{Three investigated ZSL paradigms. (a) Conventional visual-to-semantic mapping $E$ trained on classification loss. (b) Semantic autoencoder~\cite{kodirov2017semantic}, visual-to-semantic $E$ and semantic-to-visual $G$ trained on both classification and reconstruction losses. (c) The proposed SP-AEN, introducing an independent visual-to-semantic $F$ and an adversarial-style discriminator $D$ between the two subspace embeddings (blue and green triangular).}
\vspace{-0.5cm}
\label{fig:2}
\end{figure}

Zero-shot visual recognition, or more generally, zero-shot learning (ZSL), recognizes novel classes that are unseen at training stage. The community has
reached a consensus that ZSL is all about transferring knowledge from seen classes to unseen classes; Despite that there are fruitful ZSL methods, the
transfer still follows the simple but intuitive mechanism: although ``raccoon'' is unseen, we can recognize it by checking if it satisfies the ``raccoon
signature'', \eg, visual attributes ``striped tail''~\cite{farhadi2009describing, lampert2009learning, zhang2013attribute, ye2017video}, classeme
``fox-like''~\cite{torresani2010efficient, li2010object, zhang2017visual, shang2017video}, or ``raccoon'' word
vectors~\cite{pennington2014glove,mikolov2013distributed}. These attributes can be modeled at training stage and are expected to be sharable in both seen and
unseen classes at test stage. After a decade of progress, the transfer has evolved from primitive attribute classifiers~\cite{lampert2009learning} to semantic
embedding based framework~\cite{akata2013label,frome2013devise, weston2010large}, which is prevailing due to its simple and effective paradigm (cf.
Figure~\ref{fig:2} (a)): first, it maps images from visual space to semantic space where all the classes reside; then, ZSL is reduced to a simple nearest
neighbor search --- the image is assigned to the nearest class in embedding space.

The semantic transfer ability of this embedding-based ZSL framework is limited by the \textbf{semantic loss} problem. As shown in Figure~\ref{fig:1}, discarding the low-variance attributes (\ie, less discriminative) is beneficial to classification at training; However, due to the semantic discrepancy between seen and unseen classes, these attributes would be discriminative at test time, resulting in a lossy semantic space that is problematic for unseen class recognition. The main reason is that although the class embedding has rich semantic meanings, it is still a lonely point in the semantic space, where the mappings of many images will inevitably collapse to it~\cite{marcobaroni2015hubness, fu2015transductive}. One may consider the extreme case that all the class embeddings are one-hot label vectors, degenerating to the traditional supervised classification, therefore, no semantics can be transfered.

An arguably possible solution is to preserve semantics by reconstruction --- the embedded semantic vector from one image should be able to map the image back, where any two semantic embeddings are expected to preserve sufficient semantics to be apart, otherwise the reconstruction would fail~\cite{kim2017learning, yi2017dualgan, zhu2017unpaired, he2016dual}. However, reconstruction and classification are essentially two conflicting objectives: the former aims to preserve as many image details as possible while the latter focuses on suppressing irrelevant ones. For example, using only ``head'' and ``torso'' attributes might be sufficient for ``person'' recognition while the color attributes ``red'' and ``white'' are indeed disturbing. To further illustrate this, as shown in Figure~\ref{fig:2} (b), suppose $E$: $\mathcal{V}\rightarrow \mathcal{S}$ and $G$: $\mathcal{S}\rightarrow \mathcal{V}$ are two mapping transformations between the visual and semantic spaces. For classification, we want $x, x'\in \mathcal{V}$ of the same class to be mapped to close semantic embeddings $s, s'\in\mathcal{S}$, \ie, $E(x) = s \approx s' = E(x')$; For reconstruction, we want $G(s)\approx x$ and $G(s')\approx x'$, which is difficult to be satisfied as $s\approx s'$. Therefore, joint training of the two objectives is ineffective to preserve semantics (\eg, SAE~\cite{kodirov2017semantic}). For example, as illustrated in Figure~\ref{fig:1} (b), if we want to achieve good classification performance, the reconstruction will fail generally.

To resolve this conflict, we propose a novel visual-semantic embedding framework: Semantics-Preserving Adversarial Embedding Network (SP-AEN). As illustrated in Figure~\ref{fig:2} (c), we introduce a new mapping $F$: $\mathcal{V}\rightarrow \mathcal{S}$ and an adversarial objective~\cite{goodfellow2014generative} where the discriminator $D$ and encoder $F$ try to make $F(x)$ and $E(x)$ indistinguishable. There are two benefits of introducing $F$ and $D$ to help $E$ preserve semantics: 1) \textbf{Semantic Transfer}. Even though the semantic loss is inevitable by $E$, we can avoid it using $F$ by borrowing ingredients from $E(x)$ of other classes, and the discriminator $D$ will eventually transfer semantics from $F(x)$ to $E(x)$ by tailoring the two semantic embeddings into the same distribution. For example, for a ``bird'' image where the attribute ``spotty'' in $E(\textrm{bird})$ is lost, we can retain it by using $E(\textrm{leapard})$ because ``spotty'' is a discriminative and preserved attribute in ``leopard'' images. 2) \textbf{Disentangled Classification and Reconstruction}. As the reconstruction is only imposed to $F$ and $G$, $E$ is disentangled to focus on classification. In this way, the conflict between classification and reconstruction is resolved because the constraint $G(E(x)) \approx x$ and $G(E(x')) \approx x'$ is relaxed to $G(F(x))\approx x$ and $G(F(x'))\approx x'$, as $F(x)$ and $F(x')$ are not necessarily to be close with each other to comply with the discriminative objective as $E$. As shown in Figure~\ref{fig:1} (b), compared to the reconstruction style in Figure~\ref{fig:2} (b)~\cite{kodirov2017semantic}, our visual-semantic embedding $G(F(x))$ can reconstruct photo-realistic images, suggesting that the semantic is better preserved.

\begin{figure*}[htbp]
	\centering
	\includegraphics[width=1\linewidth]{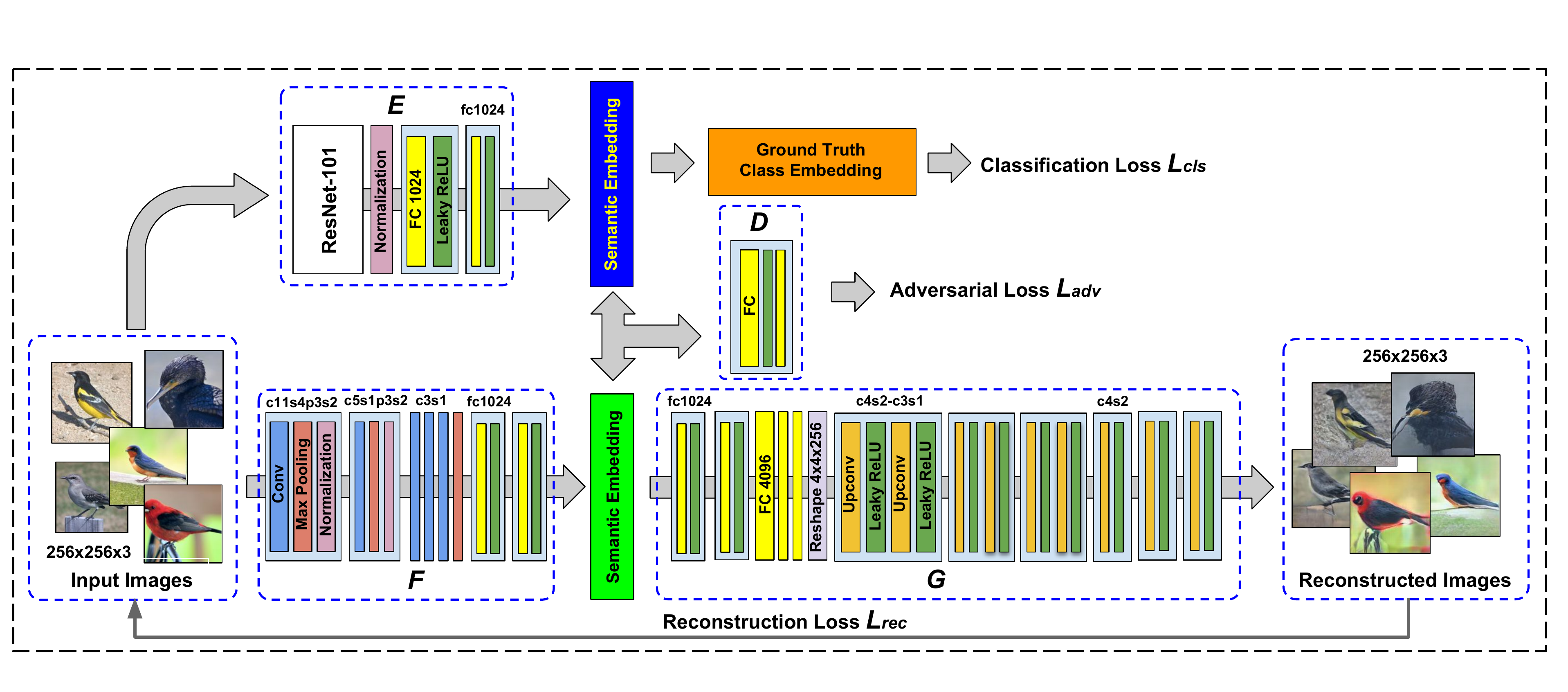}
	\caption{The architecture of SP-AEN with corresponding kernel size c, number of fully-connected layer dimension fc, and stride s of each convolutional layer. Same color indicates the same layer type.}
\vspace{-0.5cm}
\label{fig:3}
\end{figure*}

We can deploy state-of-the-art network structures for SP-AEN in a flexible plug-and-play and end-to-end fine-tune fashion, \eg, $E$ may use the powerful model for classification~\cite{he2016deep}, $F$ and $G$ may use the encoder and decoder of the image generation architecture~\cite{dosovitskiy2016generating}. The overall architecture is illustrated in Figure~\ref{fig:3} and will be detailed in Section~\ref{sec:arch}. We validate the effectiveness of SP-AEN on four popular benchmarks: CUB~\cite{welinder2010caltech}, AWA~\cite{lampert2009learning}, SUN~\cite{patterson2012sun}, and aPY~\cite{farhadi2009describing}, surpassing the state-of-the-art performances~\cite{xian2017zero} by 12.2\%, 9.3\%, 4.0\%, and 3.6\% in harmonic mean values, respectively. To the best of our knowledge, SP-AEN is the first ZSL model that empowers photo-realistic image generation from the semantic space. We hope that it will facilitate the ZSL community for better visual investigations of knowledge transfer.

\section{Related Work}
\textbf{Zero-Shot Learning}
One main stream of ZSL is the attribute-based visual recognition~\cite{farhadi2009describing,lampert2009learning,romera2015embarrassingly,norouzi2013zero, Demirel_2017_ICCV, Jiang_2017_ICCV} where the attributes serve as an intermediate feature space that transfer semantics across classes, supporting zero-shot recognition of unseen classes. To scale up ZSL, embedding based methods are prevailing~\cite{frome2013devise, akata2016label, akata2015evaluation, romera2015embarrassingly, xian2016latent, socher2013zero, kodirov2017semantic, li2017zero}. These methods directly learn a mapping from the image visual space to a semantic space, represented by semantic vectors such as word vectors~\cite{mikolov2013distributed, pennington2014glove, niu2017feaboost} or textual descriptions~\cite{lei2015predicting,elhoseiny2013write,chen2017sca}. Our proposed SP-AEN is an embedding based ZSL that exploits the ranking based classification loss as~\cite{frome2013devise}. However, to the best of our knowledge, SP-AEN is the first ZSL method that can reconstruct images from the semantic embeddings. The evaluation used in the experiments follows a similar setting for practical ZSL applications~\cite{chao2016empirical,xian2017zero}. Note that ZSL is also closely related to few-shot learning~\cite{hariharan2016low} and domain adaptation~\cite{Motiian_2017_ICCV,Busto_2017_ICCV}, where both of them assume that a small number of training images given in the test classes; however, no image is exposed to test classes at training in ZSL.

\textbf{Domain Shift and Hubness}.
Similar problems to the semantic loss have been reported in other terms. Domain shift~\cite{saenko2010adapting,fu2015transductive} is a generic problem that resides in all types of visual recognition, where the data from train and test are in different distributions. Hubness~\cite{marcobaroni2015hubness} states the phenomenon that the mapped semantic embeddings from images would be collapsed to hubs, which are near many other points without being similar to the class label in any meaningful way. We believe that semantic loss is one of the main reason for hubness, which can be alleviated by reconstruction~\cite{kim2017learning, yi2017dualgan, zhu2017unpaired, he2016dual}. In this paper, we find that jointly training~\cite{kodirov2017semantic} reconstruction and classification is not effective to preserve semantics. Another way of countering semantic loss is to learn independent attribute classifiers~\cite{morgado2017semantically}, which is not applicable when attribute annotation is unavailable.

\textbf{Generative Adversarial Network (GAN)}.
The idea of GAN~\cite{goodfellow2014generative} is to train a generator that can fool a discriminator to confuse the distributions of the generated and true samples. In theory, this max-min training procedure can lead the generator to perfectly model the data distribution. SP-AEN is similar to the GAN applied in the feature-level~\cite{odena2016conditional, tzeng2017adversarial, aae, shrivastava2016learning}. Recently, several ZSL models adopt generative model for data augmentation of unseen classes~\cite{mishra2017generative, bucher2017generating}. However, they violate the ZSL assumption that the unseen class is prohibitively seen at training.

\textbf{Image Generation}.
We seek algorithms that can generate perceptually realistic images~\cite{gatys2015texture,johnson2016perceptual,bruna2015super, ma2017pose, ma2018disentangled, sun2018natural}. Besides pixel-level loss, these methods impose feature-level reconstruction loss for preserving perceptual similarity or adversarial loss to remove unreal artifacts. However, they are based on image-to-image transformation while we requires that the reconstruction is from the semantic embedding. Our reconstruction network relates to image generation from a bottleneck layer~\cite{dosovitskiy2016generating, nguyen2016plug,zhang2016stackgan,reed2016generative}.

\section{Formulation}
We start by formalizing the ZSL task and then introduce the training objectives of the proposed SP-AEN.
\subsection{Preliminaries}
Given a set of training set $\{x_i, l_i\}$, where $x_i\in\mathcal{V}$ is an image represented in the visual space, and $l_i\in \mathcal{L}_s$ is a class label in the \emph{seen} class set, the goal of ZSL is to learn a classifier which can generalize to predict any image $x$ at test stage to its correct label, which is  not only in $\mathcal{L}_s$ but also in the \emph{unseen} class set $\mathcal{L}_u$. As summarized in~\cite{xian2017zero,lei2015predicting}, almost all types of ZSL methods can be unified into the embedding-based framework: we hope to find a visual-to-semantic mapping $E$: $\mathcal{V}\rightarrow \mathcal{S}$, where any class label $l$ is embedded as $\mathbf{y}_l\in\mathbb{R}^d$ in the semantic space $\mathcal{S}$ (\eg, an attribute space). Therefore, the predicted label $l^*$ can be obtained by following simple nearest neighbor search:
\begin{equation}\label{eq:1}
l^* = \max_{l\in\mathcal{L}}~\mathbf{y}^T_l E(x).
\end{equation}
In particular, if $l\in\mathcal{L}_u$, it is the \emph{conventional} ZSL setting; if $l \in\mathcal{L}_s\cup \mathcal{L}_u$, it is the \emph{generalized} ZSL setting, which is more practical for real applications. It is worth noting that Eq.~\eqref{eq:1} is not necessarily a linear model; in fact, as we will introduce in Section~\ref{sec:arch}, it can also be highly non-linear in nature by using deep neural networks to implement $E$.

\subsection{Classification Objective}
As label prediction in Eq.~\eqref{eq:1} is fundamentally a ranking problem, we use a large-margin based ranking loss function for classification
objective~\cite{frome2013devise,weston2010large, liu2017hierarchical}, \ie, given a training pair $(x, l)$ we want a higher dot-product similarity between
$\mathbf{y}_l$ and $E(x)$ and a lower one for any wrongly labeled pair $(x,l')$, and the similarity margin between the correct one and the wrong one should be
larger than a threshold:
\begin{equation}\label{eq:2}
L_{cls} = \sum\limits_{l\neq l'}~\max\{0, \gamma - \mathbf{y}_l^T E(x) + \mathbf{y}^T_{l'} E(x)\}.
\end{equation}
Where $\gamma >0$ is a hyperparameter for the margin. At each iteration in stochastic training, only one $l'$ is randomly selected from the unpaired labels.

As mentioned in Section~\ref{sec:intro}, the classification objective $L_{cls}$ essentially forces the semantic embedding $E(x)$ of all the images close to same ground truth label embedding $\mathbf{y}$, resulting in the semantic loss that can be tackled by using two additional objectives introduced next.

\subsection{Reconstruction Objective}
The reconstruction objective is to learn a semantic-to-visual mapping $G$: $\mathcal{S}\rightarrow \mathcal{V}$ that reconstructs a semantic embedding $s\in\mathcal{S}$ back to image such that $\|G(s)-x\|$ is small. Recall that the reconstruction in the autoencoder fashion $s = E(x)$ conflicts with the classification objective, therefore, we introduce an independent visual-to-semantic mapping $F$ for embedding reconstructive $s = F(x)$. Moreover, being different from~\cite{kodirov2017semantic} where the visual space $\mathcal{V}$ is a feature space from the output of a higher-layer in deep CNN~\cite{he2016deep, simonyan2014very}, we directly use the raw $256\times 256 \times 3$ RGB color space for image reconstruction. The reason is that the feature space from CNN is already a semantic space~\cite{zeiler2014visualizing}, which is meant to have semantic loss since its construction.

By minimizing a reconstruction objective, $F(x)$ is expected to preserve sufficient semantics so as to reconstruct images. We follow the recent progress in generating photo-realistic images~\cite{johnson2016perceptual,dosovitskiy2016generating,ledig2016photo}:
\begin{equation}\label{eq:3}
L_{rec} = L_{feat}+\lambda_p L_{pixel}.
\end{equation}
$L_{feat} = \|\phi\left(G\left(F(x)\right)\right)-\phi(x)\|^2_2$ is the feature-level (or perceptual) loss that is shown to be effective in preserving the perceptual similarity of two images, \eg, local structure details. We use the conv5 of AlexNet~\cite{krizhevsky2012imagenet} for $\phi$.
$L_{pixel} = \|G(F(x))-x\|^2_2$ is the pixel-level reconstruction loss that stabilizes the reconstruction.

\subsection{Adversarial Objective}
Yet, the disentangled semantic embeddings $E(x)$ and $F(x)$ are not interacted with each other for semantic transfer, \ie, our goal is to combine the rich semantics preserved in $F(x)$ from multiple $E(x')$ across a variety of classes. However, it is hard to hand-engineer a plausible combination rule for the dynamic $F(x)$ and $E(x)$ during training. To this end, we apply the adversarial objective~\cite{goodfellow2014generative} to encourage $E(x)$ to favor solutions that reside on the manifold of $F(x)$ that preserves semantics, by ``fooling'' a discriminator network $D$ that outputs the probabilities that $E(x')$ is as ``real'' as $F(x)$:
\begin{equation}\label{eq:adv}
L_{adv} = \mathbb{E}_{x}\left( \log D(F(x)) \right) + \mathbb{E}_{x'}\left( \log \left[1-D(E(x'))\right] \right).
\end{equation}
Where $E$ tries to minimize $L_{adv}$ against $D$ that tries to maximize it, \ie, $E^* = \argmin_E \max_D L_{adv}$.

Minimizing $L_{adv}$ is notoriously tricky due to the well-known mode collapse problem~\cite{arjovsky2017wasserstein}. In our case, the collapse may happen if similar images $x$ and $x'$, generally in the same class, dominating $L_{adv}$ by $\|F(x)-E(x')\|\approx 0$ and thus leading to failed semantic transfer across classes.
To prevent this, we followed the strategy of WGAN~\cite{arjovsky2017wasserstein}. We empirically find that this trick helps better gradient and training stability.


\subsection{Full Objective}
Combining the three objectives introduced above, our full objective of the proposed SP-AEN is:
\begin{equation}\label{eq:full}
\begin{split}
L (E, F, G, D) = L_{cls}(E) &+ \alpha L_{rec} (E, F, G) \\
&+ \beta L_{adv}(E, F, G, D).
\end{split}
\end{equation}
Where $\alpha$ and $\beta$ are trade-off parameters for different objectives. The final goal is to solve:
\begin{equation}\label{eq:full_simple}
E^* = \argmin\limits_{E, F, G}\max\limits_{D} L(E, F, G, D).
\end{equation}

As illustrated in Figure~\ref{fig:3}. By considering $F$ as the encoder and $G$ as the decoder, then the semantic embedding $F(x)$ can be considered as the bottleneck layer, regularized to match a supervised distribution $E(x)$. In this way, our SP-AEN is a supervised Adversarial Autoencoder (AAE)~\cite{aae}. Therefore, SP-AEN has the potential flexibility to reform into other ZSL frameworks, \eg, semi-supervised SP-AEN, by imposing another adversarial objective for $F(x)$ to match a prior embedding space. We leave this for our future work.

\section{Implementation}
\subsection{Architecture}\label{sec:arch}
The overall architecture is detailed in Figure~\ref{fig:3}. It is an end-to-end network with the input of raw images and ground truth class embeddings. The embedder $E$ is based on ResNet-101~\cite{he2016deep} takes a cropped $224 \times 224 \times 3$ image as input and outputs a $d$-dimensional embedding vector, which is then fed into the classification loss function in Eq.~\eqref{eq:2}. $F$ is based on AlexNet~\cite{krizhevsky2012imagenet} appended with two more fully-connected blocks that take the raw image as input and output a $d$-dimensional embedding vector, which is fed into the subsequent reconstruction network $G$. We adapt the architecture for $G$ from~\cite{dosovitskiy2016generating} who has shown impressive results for generating images from a bottleneck layer. $G$ contains five up-convolutional blocks with leaky ReLU~\cite{he2015delving} for transforming a vector into a 3-D feature map, which is eventually equal to the RGB color map. In particular, we append two fully-connected layers as the head of $G$ that takes the embedding vector as input and then output a 4,096-dimensional vector that can match the input of~\cite{dosovitskiy2016generating}. $D$ is a two-layer fully-connected layer plus a non-linear ReLU layer that takes the $d$-dimensional embedding vector as input.

\subsection{Training Details}\label{sec:train}
For all the experiments in this paper, the training images are resized with its shorter side to 256; ten $224 \times 224$ cropped image data augmentation trick is adopted with per-pixel mean subtraction~\cite{krizhevsky2012imagenet}. For efficiency, we fixed the ResNet-101 in $E$, and initialized the AlexNet-like blocks in $F$ with AlexNet and $G$ with the pretrained generator~\cite{dosovitskiy2016generating}. Then, the rest of the modules was trained from scratch with MSRA random initializer~\cite{he2015delving}. The learning rate started from $1e^{-4}$ and is multiplied by 0.1 when the error is plateaus. We use grid search to select parameter $\alpha$ and $\beta$.

\begin{table*}[htbp]
\centering
\scalebox{.7}{
\begin{tabular}{l c c c c|c c c c|c c c c|c c c c}
\hline
\multicolumn{1}{c}{} & \multicolumn{4}{c}{\textbf{SUN} (0.9851)} & \multicolumn{4}{c}{\textbf{CUB}  (0.9575)} & \multicolumn{4}{c}{\textbf{AWA} (0.7459)} & \multicolumn{4}{c}{\textbf{aPY} (0.5847)} \\
\hline
Setting  & U$\to$U & U$\to$T & S$\to$T & H & U$\to$U & U$\to$T & S$\to$T & H & U$\to$U & U$\to$T & S$\to$T & H  & U$\to$U & U$\to$T & S$\to$T & H \\
\hline
DAP~\cite{lampert2009learning} & 39.9 & 4.2 & 25.1 & 7.2 & 40.0 &  1.7 & 67.9 & 3.3 & 46.1 & 0.0 & 84.7 & 0.0 & 33.8  & 4.8 & 78.3 & 9.0 \\
IAP~\cite{lampert2009learning} & 19.4 & 1.0 & 37.8 & 1.8 & 24.0 & 0.2 &\textbf{72.8} & 0.4 & 35.9 & 0.9 & 87.6 & 1.8 & 36.6 & 5.7 & 65.6 & 10.4 \\
SSE~\cite{zhang2015zero} & 51.5 & 2.1 & 36.4 & 4.0 & 43.9 & 8.5 & 46.9 & 14.4 & 61.0 & 8.1 & 82.5 & 14.8 & 34.0 & 0.2 & 78.9 & 0.4 \\
CONSE~\cite{norouzi2013zero} & 38.8 & 6.8 & \textbf{39.9} & 11.6 & 34.3 & 1.6 & 72.2 & 3.1 & 44.5 & 0.5 & 90.6 & 1.0 & 26.9 & 0.0 & \textbf{91.2} & 0.0 \\
SYNC~\cite{changpinyo2016synthesized} & 56.3 & 7.9 & 43.3 & 13.4 & \textbf{55.6} & 11.5 & 70.9 & 19.8 & 46.6 & 10.0 & 90.5 & 18.0 & 23.9 & 7.4 & 66.3 & 13.3 \\
\hline
CMT~\cite{socher2013zero} & 39.9 & 8.1 & 21.8 & 11.8 & 34.6 & 7.2 & 49.8 & 12.6 & 37.9 & 0.5 &  90.0 & 1.0 & 28.0 & 1.4 & 85.2 & 2.8 \\
CMT${}^\star$~\cite{socher2013zero} & --- & 8.7 & 28.0 & 13.3 & --- & 4.7 & 60.1 & 8.7 & --- & 8.7 & 89.0 & 15.9 & --- & 10.9 & 74.2 & 19.0 \\
LATEM~\cite{xian2016latent} & 55.3 & 14.7 & 28.8 & 19.5 & 49.3 & 15.2 & 57.3 & 24.0 & 55.8 & 11.5 & 77.3 & 20.0 & 35.2 & 0.1 & 73.0 & 0.2 \\
DeViSE~\cite{frome2013devise} & 56.5 & 16.9 & 27.4 & 20.9 & 52.0 & 23.8 & 53.0 & 32.8 & 59.7 & 17.1 & 74.7 & 27.8 & \textbf{39.8} & 4.9 & 76.9 & 9.2 \\
ALE~\cite{akata2016label} & 58.1 & 21.8 & 33.1 & 26.3 & 54.9 & 23.7 & 62.8 & 34.4 & \textbf{62.5} & 14.0 & 81.8 & 23.9 & 39.7 & 4.6 & 73.7 & 8.7 \\
SJE~\cite{akata2015evaluation} & 53.7 & 14.7 & 30.5 & 19.8 & 53.9 & 23.5 & 59.2 & 33.6 & 61.9 & 8.0 & 73.9 & 14.4 & 32.9 & 3.7 & 55.7 & 6.9 \\
ESZSL~\cite{romera2015embarrassingly} & 54.5 & 11.0 & 27.9 & 15.8 & 53.9 & 12.6 & 63.8 & 21.0 & 58.6 & 5.9 & 77.8 & 11.0 & 38.3 & 2.4 & 70.1 & 4.6 \\
SAE~\cite{kodirov2017semantic} & 40.3 & 8.8 & 18.0 & 11.8 & 33.3 & 7.8 & 54.0 & 13.6 & 54.1 & 1.1 & 82.2 & 2.2 & 8.3 & 0.4 & 80.9 & 0.9 \\
\hline
SP-AEN &\textbf{59.2} &  \textbf{24.9} & 38.6& \textbf{30.3} & 55.4 & \textbf{34.7} & 70.6  & \textbf{46.6} & 58.5 & \textbf{23.3} & \textbf{90.9} & \textbf{37.1} & 24.1 & \textbf{13.7} & 63.4 & \textbf{22.6} \\
\hline
\end{tabular}
}
\caption{Performances (accuracy\% and H\%)  of all the comparing methods under the three settings on the four datasets. Cosine similarity between the attribute variances of the disjoint train images and test images are given in brackets. As demonstrated in Figure~\ref{fig:1}, lower similarity indicates larger semantic loss.}
\vspace{-0.5cm}
\label{tab:1}
\end{table*}

\section{Experiments}
\subsection{Datasets}
We used four popular benchmarks described as below . In particular, we followed the new split provided by~\cite{xian2017zero} as the ILSVRC~\cite{russakovsky2015imagenet} 1K ImageNet classes, widely used as a pre-training source for CNN features, have already included the test classes in the conventional split of the benchmarks, hence violating the fundamental assumption of ZSL that the classes at test should be strictly unseen at training.
\\
\textbf{CUB}~\cite{welinder2010caltech}. It is the Caltech-UCSD-Birds 200-2011 dataset of 11,788 bird images from 200 fine-grained classes. Each image is annotated with 312 semantic attributes. The train split has 7,057 images across 150 classes (50 classes for validation); The test split has 1,764 images from the 150 seen classes and 2,967 images from the 50 unseen classes.
\\
\textbf{SUN}~\cite{patterson2012sun}. It is a fine-grained scene data of 14,340 images across 717 scene classes. Each image is annotated with 102 semantic attributes. The train split has 10,320 images from 645 classes (65 classes for validation); The test split has 2,580 images from the 645 seen classes and 1,440 images from the 72 unseen classes.
\\
\textbf{AWA}~\cite{lampert2009learning}. It is the coarse-grained Animals with Attributes dataset of 30,475 images from 50 animals. Each class is annotated with 85 semantic attributes. The train split has 23, 527 images from 40 classes (13 classes for validation); The test split has 5,882 images from the 40 seen classes and 7,913 images from the 10 unseen classes. In particular, we used the AWA2 released by~\cite{xian2017zero} as the images from the original one are restricted due to photo copyright reasons.
\\
\textbf{aPY}~\cite{farhadi2009describing}. It is the coarse-grained Attribute Pascal and Yahoo dataset of 12,051 images from 32 generic object classes (\eg, 20 Pascal classes and 12 popular Yahoo classes). Each image is annotated with 64 semantic attributes. The train split has 5,932 images from 20 classes (5 classes for validation); The test split has 1,483 images from the 20 seen classes and 7,924 from the 12 unseen classes.

For fair comparison and reproductivity,  we used the class label embeddings provided by~\cite{xian2017zero}, each of which is an L2-normalized vector.
\subsection{Settings and Evaluation Metrics}
To evaluate the ZSL performances over all classes, we applied the following three settings. 1) U$\to$ U: The test images and the prediction labels set are limited to the unseen classes; 2) S$\rightarrow$ T: The test images are from the seen classes and the prediction labels set is the union of both seen and unseen classes; 3) U$\rightarrow$T: The test images are from the unseen classes and the prediction labels set is the union of both seen and unseen classes. Note that U$\rightarrow$ U and U$\rightarrow$ T are also known as the conventional and the generalized ZSL settings.

We followed~\cite{xian2017zero} and used the per-class top-1 accuracy as the evaluation metric, where the prediction using Eq.~\eqref{eq:1} is successful if the predicted class is the correct ground truth. We averaged the accuracies of all classes. For generalized ZSL setting, we also used the recently proposed harmonic mean ($H$)~\cite{xian2017zero} of accuracies on seen classes $\mathcal{L}_s$ ($Acc_{S \rightarrow T}$) and unseen classes $\mathcal{L}_u$ ($Acc_{U \rightarrow T}$) :
\begin{equation}\label{eq:H}
H = 2\times Acc_{S\rightarrow T}\times Acc_{U\rightarrow T} /(Acc_{S\rightarrow T}+Acc_{U\rightarrow T}).
\end{equation}
$H$ offers a comprehensive metric in the practical ZSL case: in many real applications, the test image would belong to any class from both seen and unseen sets, and it is required that the accuracy should be high on both of them. Note that we slightly abuse the setting notations: U$\to$ U, S$\to$ T, U$\to$ T as the accuracy calculated in the corresponding setting.

\begin{figure*}[htbp]
	\centering
	\includegraphics[width=1.0\linewidth]{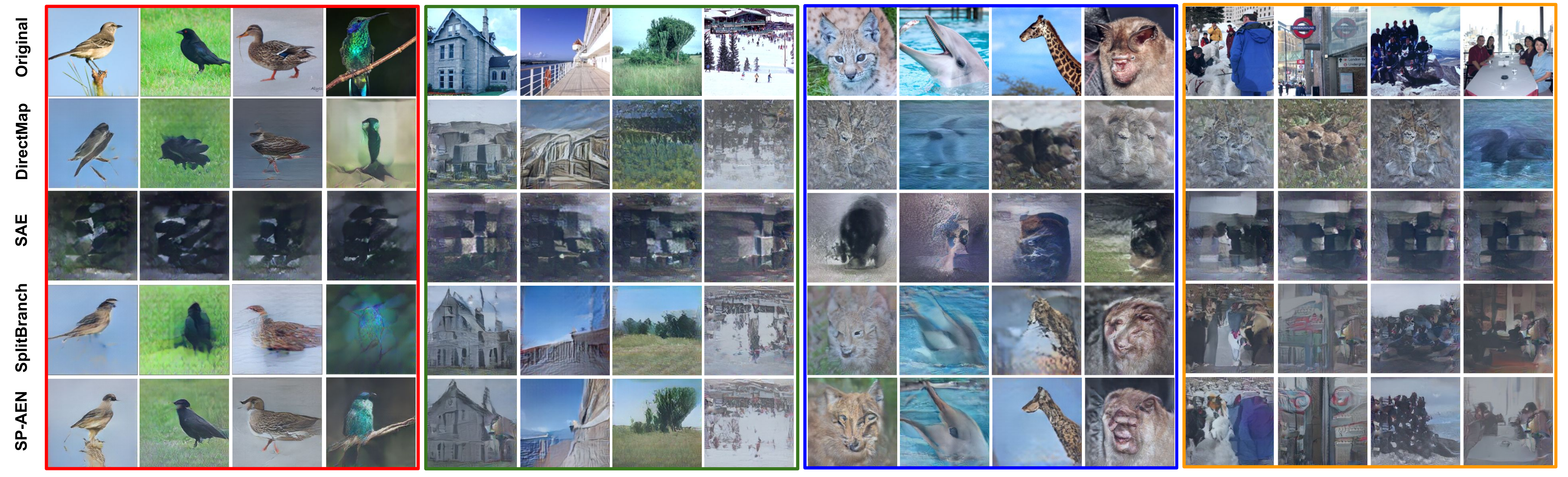}
	\caption{Example reconstruction results of various architectures on CUB, SUN, AWA and aPY respectively.}
\label{fig:5}
\end{figure*}

\begin{table*}[htbp]
\centering
\scalebox{.7}{
\begin{tabular}{l c c c c|c c c c|c c c c|c c c c}
\hline
\multicolumn{1}{c}{} & \multicolumn{4}{c}{\textbf{SUN}} & \multicolumn{4}{c}{\textbf{CUB}} & \multicolumn{4}{c}{\textbf{AWA}} & \multicolumn{4}{c}{\textbf{aPY}} \\
\hline
Setting  & U$\to$U & U$\to$T & S$\to$T & H & U$\to$U & U$\to$T & S$\to$T & H & U$\to$U & U$\to$T & S$\to$T & H  & U$\to$U & U$\to$T & S$\to$T & H \\
\hline
Cls. Only & 56.8 & 17.2 & 29.0 & 21.6 & 52.2  & 23.5 & 55.0 &32.9 & \textbf{60.2} &17.5 & 76.7& 28.5 & \textbf{35.8} & 5.5 & \textbf{72.9} & 10.2\\
Full Obj. &\textbf{59.2} &  \textbf{24.9} & \textbf{38.6}& \textbf{30.3} & \textbf{55.4} & \textbf{34.7} & \textbf{70.6}  & \textbf{46.6} & 58.5 & \textbf{23.3} & \textbf{90.9} & \textbf{37.1} & 24.1 & \textbf{13.7} & 63.4 & \textbf{22.6} \\
\hline
\end{tabular}
}
\caption{Performances (accuracy\% and H\%)  of all the comparing methods under the three settings on the four datasets.
}
\vspace{-0.5cm}
\label{tab:3}
\end{table*}

\subsection{Comparisons with State-of-The-Arts}
\textbf{Comparing Methods}. We compared SP-AEN with a variety of ZSL methods as reported in~\cite{xian2017zero}. These methods fall into two categories. 1)
\emph{embedding based}: DeViSE~\cite{frome2013devise}, ALE~\cite{akata2016label}, SJE~\cite{akata2015evaluation}, ESZSL~\cite{romera2015embarrassingly},
LATEM\cite{xian2016latent}, CMT/CMT$^*$~\cite{socher2013zero}, SAE~\cite{kodirov2017semantic}. As SP-AEN does, this category maps images into the semantic
embedding space where all class labels reside. Note that CMT${}^\star$ is CMT with novelty detection and hence is not applicable for U$\to$U setting. To the
best of our knowledge, SAE is the only ZSL method that uses reconstruction to tackle the semantic loss problem. 2) \emph{attribute based}:
DAP~\cite{lampert2009learning}, IAP~\cite{lampert2009learning}, SSE~\cite{zhang2015zero}, CONSE~\cite{norouzi2013zero}, and
SYNC~\cite{changpinyo2016synthesized}. These methods are based on an intermediate inference of attributes in ZSL. Note that this category cannot be applied in
generic class embeddings without the attribute annotations.

\textbf{Results}. Table~\ref{tab:1} summarizes the performances (accuracy\% and H\%)  of all the comparing methods under the three settings on the four
datasets. We have the following two key observations: 1) Using the generalized ZSL setting metric (U$\to$T and H), SP-AEN significantly outperforms the best
competitors by around 4\% to 12\%. In particular, we can clearly see that the performance gap between SP-AEN and others becomes larger as the cosine
similarity between the attribute variances of the disjoint train and test splits. As larger cosine similarity indicates smaller semantic loss, it demonstrates
the effectiveness of SP-AEN in alleviating the semantic loss of ZSL. 2) Under the conventional ZSL setting (U$\to$U), in most cases, SP-AEN achieves the best
performance. This is reasonable as the search space of label prediction is merely limited to the unseen sets, however, the semantic loss may cause the
semantic mapping of unseen class images very similar to one of the seen classes, resulting in incorrect recognition.

\begin{figure*}[htbp]
	\centering
	\includegraphics[width=1\linewidth]{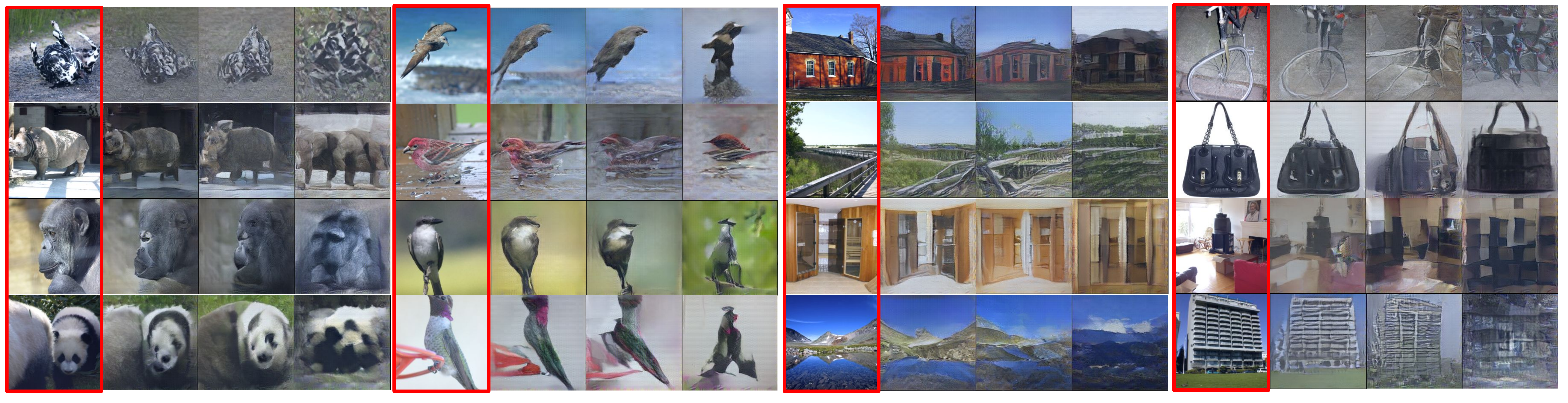}
	\caption{Reconstruction results of different $\alpha$ in AWA, CUB, SUN and aPY respectively. The left image in the red bounding box represents the original images, and decreases weights $\alpha$ in Eq.~\ref{eq:full} from left to right.}
\vspace{-0.5cm}
\label{fig:6}
\end{figure*}

\begin{table}[htbp]
\centering
\scalebox{.8}{
\begin{tabular}{c c| c| c| c}
\hline
Method & \textbf{SUN} & \textbf{CUB} & \textbf{AWA} & \textbf{aPY} \\
\hline
 DirectMap & 0.079 & 0.069 & 0.075 & 0.085 \\
 SAE & 0.285 & 0.281 & 0.259 &  0.275\\
SplitBranch & 0.070 & 0.058  & 0.059 & 0.076 \\
SP-AEN & \textbf{0.053}  & \textbf{0.040}& \textbf{0.047} & \textbf{0.055} \\
\hline
\end{tabular}
}
\caption{The mean squared pixel-level loss between the input images and its reconstructed images of various reconstruction settings over four datasets.}
\vspace{-0.5cm}
\label{tab:2}
\end{table}

\subsection{Ablation Studies}
\subsubsection{Conflict between Classification \& Reconstruction}
\begin{figure}[t]
	\centering
	\includegraphics[width=0.7\linewidth]{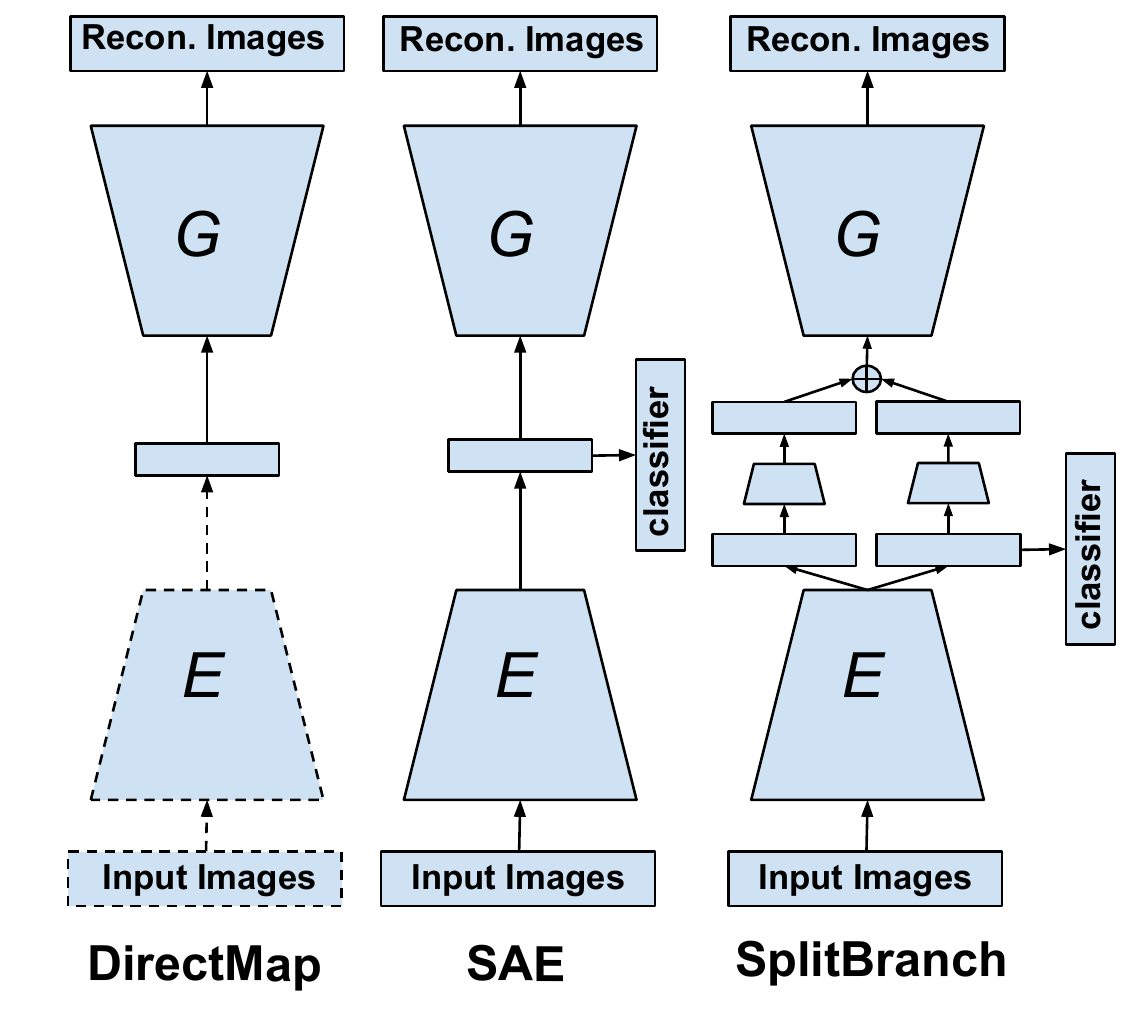}
	\caption{Three reconstruction architectures evaluated. Dashed line indicates fixed parameters at training.}
\vspace{-0.5cm}
\label{fig:4}
\end{figure}

To validate our key motivation for the design of SP-AEN: The conflict between classification and reconstruction, as illustrated in Figure~\ref{fig:4}. We propose three possible architectures that can achieve the semantic-to-visual reconstruction as SP-AEN: 1) \textbf{DirectMap}: For each input image, we use $E$ to get its discriminative semantic embedding and then use $G$ to map it back to the image space. In this architecture, we fix $E$ and train $G$. DirectMap is used to evaluate how much reconstructive semantics are preserved in the discriminatively trained semantic embedding. 2) \textbf{SAE}~\cite{kodirov2017semantic}: We adapt the SAE architecture using our image reconstructor $G$ as the decoder and $E$ as the encoder. The bottleneck layer as semantic embedding is used for classification. We jointly train \textit{E} and \textit{G}. 3) \textbf{SplitBranch}: We split the semantic output of encoder \textit{E} into two branches, and only the first semantic embedding branch is used for classification. Two semantic embedding branches are concatenated after two respective fully-connected layers. The merged semantic representation is fed to decoder \textit{G} to reconstruct image.

\textbf{Reconstruction Results}. Figure~\ref{fig:5} shows some reconstructed images and Table~\ref{tab:2} reports the reconstruction losses of the unseen images in the test splits of the four datasets. We can have the following observations:
1) On CUB and SUN, the images reconstructed by DirectMap are close to SP-AEN, which have the highest quality. However, on AWA and aPY, the reconstruction quality of DirectMap significantly drops. Again, this is due to the semantic loss between train and test set, as the attribute variance cosine similarities of AWA and aPY are much larger than those of CUB and SUN.
2) If we jointly train the discriminative $E$ and reconstructive $G$ as in SAE, the reconstruction fails in all examples; If we jointly train them as in SplitBranch, we can observe significant quality improvement, closing to SP-AEN. However, we find that the weight for merging the semantic embedding of the classification branch is almost zero, meaning the contribution of the semantic embedding for reconstruction is minor. Therefore, the semantic transfer is ineffective. This motivates us to use adversarial loss in SP-AEN to allow semantic transfer and high-quality reconstruction at the same time.


\subsubsection{Effectiveness of $D$ and $G$}
Since the score of seen classes is usually larger than that of unseen classes, a calibrated stacking rule~\cite{chao2016empirical} to solve it by subtracting a bias for seen classes to solve it:
$$ l^* = \max_{l\in \mathcal{L}_u \cup \mathcal{L}_s }~\mathbf{y}^T_l E(x) - \gamma \mathbbm{1} \left[ l \in \mathcal{L}_s \right]. $$
Where the indicator function $\mathbbm{1} \left[ \cdot \right]$  indicates whether or not $l$ is a seen class and $\gamma \in \mathbb{R}$ is a calibration factor. This calibrated stacking rule represents a middle ground between aggressively classifying each data point into seen classes and conservatively classifying every data point into unseen classes. By varying the calibration factor $\gamma$, we can compute a series of classification accuracies ($Acc_{U \to T}$ and $Acc_{S \to T}$) and plot the Seen-Unseen accuracy Curve (SUC). The Area Under Seen-Unseen Accuracy Curve (AUSUC) is always used as a performance metric to show the balance capability between this two conflicting objectives $Acc_{U \to T}$ and $Acc_{S \to T}$ in the generalized ZSL.

\begin{figure}[htbp]
	\centering
	\includegraphics[width=1\linewidth]{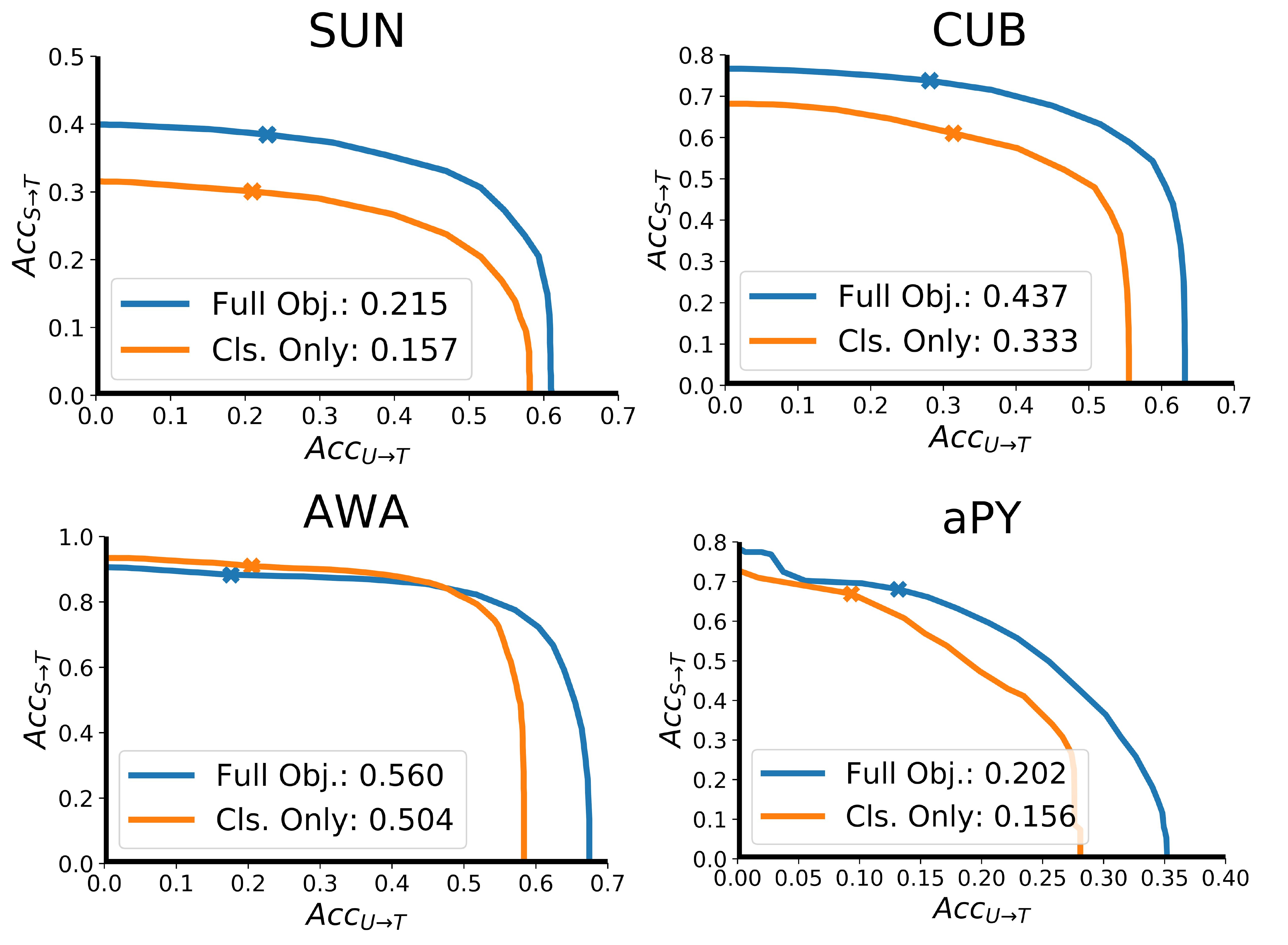}
	\caption{The Seen-Unseen accuracy Curve (SUC)~\cite{chao2016empirical} of SUN, CUB, AWA and aPY respectively. The blue line represents SP-AEN with whole loss objectives and the orange line represents SP-AEN with only classification loss objective. The cross denotes direct stacking when calibration factor $\gamma = 0$.}
\label{fig:ausuc}
\end{figure}

Table~\ref{tab:3} reports ablative results of SP-AEN without the reconstruction $G$ and discriminator $D$ (Cls. Only)
. We can observe that by using adversarial training, we can significantly improve the $H$ value by over 10\% on all datasets. Figure~\ref{fig:ausuc} shows the AUSUC~\cite{chao2016empirical} of SP-AEN model with full objective (Full Obj.) and with only classification objective (Cls. Only)\footnote{The performance results on Table~\ref{tab:3} and Figure~\ref{fig:ausuc} are based on different hyper parameters setting. For results in Table~\ref{tab:3}, $\alpha$ and $\beta$ are set to 10 and 5, and for results in Figure~\ref{fig:ausuc}, $\alpha$ and $\beta$ are set to 10 and 50.}. We can observe that SP-AEN (Full Obj.) is consistently larger than model(Cls. Only) over all datasets. Both two types of metrics consistently demonstrates that SP-AEN enables effective semantic transfer. Figure~\ref{fig:6} illustrates shows that by lowering the trade-off of $G$ compared to $D$, the reconstruction quality will reduce.

\section{Conclusions}
We proposed a novel embedding based ZSL framework called Semantics-Preserving Adversarial Embedding Network (SP-AEN) to tackle the semantic loss problem in ZSL, which was rarely addressed by prior works. SP-AEN solves this problem by a novel visual reconstruction paradigm: 1) Introducing an independent visual-to-semantic mapping and then the reconstruction from the semantic space to the visual space would not affect the classification objective, whose contradiction to the reconstruction objective is extensively validated in this paper. 2) Semantic transfer can be achieved by adversarial learning between the two independent semantic embeddings. The first step preserves the semantics via reconstruction, while the second step enables semantics transfer across classes. We validated the effectiveness of SP-AEN through extensive comparative and ablative experiments on four ZSL benchmarks.

Our future works may focus on 1) incorporating generative models into SP-AEN so as to hallucinating photo-realistic images for unseen or even synthesized classes, and 2) developing new ZSL frameworks such as semi-supervised SP-AEN by imposing a prior semantic space.

\section*{Acknowledgement}
This work was supported by National Key Research and Development Program of China (2017YFB0203001), National Natural Science Foundation of China (61572431),
Zhejiang Natural Science Foundation (LZ17F020001), Key R\&D Program of Zhejiang Province (2018C01006) and Joint Research Program of ZJU \& Hikvision Research
Institute.

{
\small
\bibliographystyle{ieee}
\bibliography{egbib}
}

\end{document}